# Automated reasoning using possibilistic logic : semantics, belief revision and variable certainty weights

Didier DUBOIS, Jérôme LANG, Henri PRADE

*Laboratoire Langages et Systèmes Informatiques*
*Institut de Recherche en Informatique de Toulouse*
*Université Paul Sabatier, 118 route de Narbonne*
*31062 TOULOUSE Cedex (FRANCE)*

**Abstract :** In this paper an approach to automated deduction under uncertainty, based on possibilistic logic, is proposed ; for that purpose we deal with clauses weighted by a degree which is a lower bound of a necessity or a possibility measure, according to the nature of the uncertainty. Two resolution rules are used for coping with the different situations, and the refutation method can be generalized. Besides the lower bounds are allowed to be functions of variables involved in the clause, which gives hypothetical reasoning capabilities. The relation between our approach and the idea of minimizing abnormality is briefly discussed. In case where only lower bounds of necessity measures are involved, a semantics is proposed, in which the completeness of the extended resolution principle is proved. Moreover deduction from a partially inconsistent knowledge base can be managed in this approach and displays some form of non-monotonicity.

## 1. Introduction

Several approaches have been recently proposed for automated theorem proving under uncertainty and/or vagueness. Some of them (Shen et al. [21], Orci [19], Martin et al. [17], Ishizuka and Kanai [14]) are more or less based on Lee's resolution method for vague predicates [16] which is also applied without caution to uncertainty degrees attached to ordinary predicates. Hinde [13] and Umano [22] use fuzzy truth-values but do not give any axiomatic basis to the underlying logic and cannot manage the use of certainty degrees. Shapiro [20] gives a more general framework for theorem proving with uncertainties. Van Emden [23] uses such a model with a MYCIN-like propagation of uncertainty degrees in the resolution process. Baldwin [1] uses formulas valued by a pair of lower and upper probabilities; some other approaches are based on probabilistic logic (Nilsson [18], Grosof [12]). Our approach (see Dubois, Prade [6], [7]) ; Dubois, Lang, Prade [4]) is based on an extension of the resolution principle in possibilistic logic, where we handle clauses weighted by a degree which is a lower bound of a necessity or a possibility measure. Thus the uncertainty degrees have a clear meaning in the framework of incomplete information systems as recalled in the next section and a distinction is made as in modal logic between what is (somewhat) certain and what is only (more or less) possible. Here we show how our approach can manage gradual rules such as "the truer $P(x)$, the more certain (or possible) $Q(x)$", where P is a vague predicate which can be satisfied to an intermediary degree and Q is an ordinary predicate, (e.g. "the younger the person, the more certain he/she is a single"). We make a strong distinction between degrees of uncertainty due to a state of incomplete knowledge, and intermediary degrees of truth due to the presence of vague predicates (see Dubois, Prade [8]). Then we will discuss the links between possibilistic reasoning and minimization of abnormality. Lastly, we will define a semantics for possibilistic logic and we show how, by defining inconsistency as a gradual notion, it enables us to make a non-trivial use of partially inconsistent knowledge bases, displaying some form of non-monotonicity.

## 2. Background

Possibilistic logic manipulates propositional calculus formulas or first-order logic closed formulas, to which a possibility degree or a necessity degree, between 0 and 1, is attached. Let be $\Pi(p)$ (respectively $N(p)$) the possibility (resp. necessity) degree of p. We adopt the following relations and conventions :

- $\forall p, \Pi(p) = 1 - N(\neg p)$, i.e. to say that $\neg p$ is all the more certainly true as p is more impossible and conversely. This is the numerical counterpart of a well-known relation in modal logic.
- $N(p)=1$ means that, given the available knowledge, p is certainly true; $\Pi(p)=0$, that it is impossible for p to be true. This contrasts with
- $\Pi(p) = \Pi(\neg p) = 1$ (equivalent to $N(p) = N(\neg p) = 0$) which expresses that, from the available knowledge, nothing can infirm nor confirm p (this is the case of total ignorance).
- $\Pi(\emptyset) = N(\emptyset) = 0$; $\Pi(\mathbf{1}) = N(\mathbf{1}) = 1$ where $\emptyset$ and $\mathbf{1}$ denote the contradiction and the tautology respectively.
- $\forall p, \forall q, \Pi(p \vee q) = \max(\Pi(p), \Pi(q))$. This is the basic axiom of possibility measures (Zadeh[24] ; Dubois,

Prade [5]), which supposes that the imprecise or vague knowledge upon which the attribution of possibility or necessity degrees is based, can be described in terms of fuzzy sets (or equivalently in terms of a nested family of ordinary subsets); see [5]. This is equivalent to $N(p \wedge q) = \min(N(p), N(q))$. However we only have $\Pi(p \wedge q) \leq \min(\Pi(p), \Pi(q))$ (no equality in the general case), and similarly, we have only $N(p \vee q) \geq \max(N(p), N(q))$.

Besides, we have $\max(\Pi(p), \Pi(\neg p)) = 1$ for any classical formula p, i.e. which does not involve any vague predicate or vague quantifier, (indeed $p \vee \neg p = \mathbf{1}$ in this case) ; this implies that $N(p) > 0 \Longrightarrow \Pi(p) = 1$ (i.e. $N(\neg p) = 0$), which means that a formula is completely possible before being somewhat certain.

This contrasts with fuzzy logic, as in Lee [16], where the degree of truth v satisfies both $v(\tilde{p} \wedge \tilde{q}) = \min(v(\tilde{p}), v(\tilde{q}))$ and $v(\tilde{p} \vee \tilde{q}) = \max(v(\tilde{p}), v(\tilde{q}))$ as well as $v(\tilde{p}) = 1 - v(\neg \tilde{p})$ for any vague propositions $\tilde{p}$ and $\tilde{q}$. Note that these vague propositions no longer obey excluded-middle or contradiction laws.

Possibilistic logic is well-adapted to the representation of states of incomplete knowledge, since we can distinguish between the complete lack of certainty in the falsity of a proposition p ($N(\neg p) = 0$) and the total certainty that p is true ($N(p) = 1$). $N(p) = 1$ entails $N(\neg p) = 0$ but the converse is false. It contrasts with probability measures where $\text{Prob}(p) = 1$ is equivalent to $\text{Prob}(\neg p) = 0$.

## 3. Possibilistic resolution

Lee [16] proposed a deduction method in fuzzy logic, which is a generalization of the resolution principle; the resolution rules we present here are not in the framework of fuzzy logic, but in possibilistic logic ; the latter, closer to classical logic than fuzzy logic, enables us to represent knowledge whose truth or falsity is uncertain, but whose content is not vague. The classical rule for propositional clauses is generalized by

(1) $$\frac{N(p \vee q) \geq \alpha \quad N(\neg p \vee r) \geq \beta}{N(q \vee r) \geq \min(\alpha, \beta)}$$

in case of lower bounds on necessity measures (Dubois, Prade [6]) ; and the particularization rule is extended by

(2) $$\frac{N(\forall x\, p(x)) \geq \alpha}{N(p(a)) \geq \alpha}$$

This pattern holds for any substitution applied to a clause. Besides the following rule (Dubois, Prade [7])

(3) $$\frac{N(p \vee q) \geq \alpha \quad \Pi(\neg p \vee r) \geq \beta}{\Pi(q \vee r) \geq \alpha \mid_m \beta},$$

where $\alpha \mid_m \beta = \begin{cases} \beta & \text{if } \alpha + \beta > 1 \\ 0 & \text{if } \alpha + \beta \leq 1, \end{cases}$

holds when one of the lower bounds qualifies a possibility measure, as well as the counterpart of (2)

(4) $$\frac{\Pi(\forall x\, p(x)) \geq \alpha}{\Pi(p(a)) \geq \alpha}$$

It can be shown that the lower bounds obtained by these rules are the best possible ones. Besides the similarity between (1)-(3) and resolution patterns existing in modal logics has been pointed out [7]; the reader is referred to [10] for a preliminary study of the links between possibility theory and modal logic.

An uncertain clause is a first-order logic clause to which a valuation is attached ; it is a lower bound of its necessity or possibility measure. Thus, in the following we shall write (c $(N\ \alpha)$) (resp. (c $(\Pi\ \alpha)$)) as soon as the inequality $N(c) \geq \alpha$ (resp. $\Pi(c) \geq \alpha$) is known. Since $N(p) > 0$ implies $\Pi(p) = 1$, it is sufficient to consider clauses which are weighted either in terms of necessity or in terms of possibility. Possibilistic resolution only handles conjunctions of uncertain clauses. However, it can also handle general formulas valued by a lower bound of a necessity measure, since the formulas can be put in conjunctive normal form and we can apply $N(p \wedge q) = \min(N(p), N(q))$.

The refutation method can be extended to possibilistic logic (Dubois, Prade [6], [7]). Indeed if we are interested in proving that p is true, necessarily or possibly to some degree, we add in the knowledge base $\mathcal{H}$ the assumption

$$N(\neg p) = 1$$

i.e. that p is false (with total certainty). Let $\mathcal{H}'$ be the new knowledge base. Then it can be proved (see [6],[7]) that any valuation attached to the empty clause produced by extended resolution using patterns (1)-(4) from $\mathcal{H}'$ is a lower bound $\alpha$ of the necessity (resp. possibility) measure of the conclusion p, if its form is $(N\ \alpha)$ (resp. $(\Pi\ \alpha)$). It entails the existence of "optimal refutations", i.e. derivations of an empty clause with a maximal valuation, the valuations being ordered by

$(N\ \alpha) \leq (N\ \beta)$ if and only if $\alpha \leq \beta$
$(\Pi\ \alpha) \leq (\Pi\ \beta)$ if and only if $\alpha \leq \beta$
$(\Pi\ \alpha) \leq (N\ \beta)$ for any $(\alpha, \beta) \in [0,1[ \times ]0,1]$.

In order to find an optimal refutation from a set of weighted clauses without expanding a too large number of nodes, resolution strategies have been proposed (see Dubois, Lang, Prade [4]).

*N.B. : Possibilistic reasoning and the minimization of abnormality.*

Possibilistic reasoning is in agreement with the idea of minimizing abnormality in commensense reasoning: a clause like $(\neg p(x) \vee q(x) \, (N \, \alpha))$, once instanciated with a particular x, say a, means that there is a possibility at most equal to $1-\alpha$ (i.e. $\Pi(p(a) \wedge \neg q(a)) \leq 1-\alpha$) that this particular x is an exception of the rule "if x satisfies p, then it satisfies q". Another way of handling a rule with (potential) exceptions is to introduce an abnormality predicate, say "ab", specific of the rule, and to state the totally certain rule $(\neg p(x) \vee q(x) \vee ab(x) \, (N \, 1))$ and to add to the knowledge base the default assumption $(\neg ab(x) \, (N \, \alpha))$, i.e., we are at least certain at a degree $\alpha$ that a given x is not a priori abnormal. Then the search for the largest weight attached to an empty clause derived from a set of clauses without abnormality predicates corresponds to try to obtain the empty clause using only the most certain clauses of the form $(\neg ab_i(x) \, (N \, \alpha_i))$, now i.e. minimizing the abnormality. For example, let $\mathcal{H}$ be the knowledge base

C1 $\neg bird(x) \vee flies(x) \vee ab1(x) \, (N \, 1)$
C2 $\neg lives\text{-}in\text{-}Antarctica(x) \vee \neg bird(x) \vee \neg flies(x) \vee ab2(x) \, (N \, 1)$
C3 $\neg ab1(x) \, (N \, 0.8)$
C4 $\neg ab2(x) \, (N \, 0.9)$

The abnormality predicate ab1 will be true for each bird which does not fly, while ab2 will be true for each bird living in Antarctica which flies. If we add bird(Tweety) to the knowledge base, then using C1 and the abnormality predicate ab1, we deduce flies(Tweety) with the certainty degree 0.8. Then, if we add bird(Tweety) and lives-in-Antarctica(Tweety), then we have to choose between using C3 and using C4 ; then we minimize the abnormality, i.e. we choose the most certain non-abnormality clause, which is C4, and then we deduce $\neg$flies(Tweety) with the certainty degree 0.9. Note that the levels of uncertainty introduce some priority between the clauses; if we want to give priority to the most specific rule about birds living in Antarctica (when it applies), the exceptions to this rule should be more abnormal than the exceptions to the rule concerning general birds (as it is the case in our example). These abnormality predicates are reminiscent of those used in circumscription, e.g., [15].

## 4. Semantic aspects

A semantics has been defined for clauses weighted by lower bounds of a necessity measure. If p is a closed formula, M(p) the set of the models of p, then the models of $(p \, (N \, \alpha))$ will be defined by a fuzzy set $M(p \, (N \, \alpha))$ with a membership function

$\mu_{M(p \, (N \, \alpha))}(I) = 1$ if $I \in M(p)$
$\quad\quad\quad\quad\quad\quad\quad = 1 - \alpha$ if $I \in M(\neg p)$.

Then the fuzzy set of the models of a knowledge base $\mathcal{H} = \{C_1, C_2, ..., C_n\}$, where $C_i$ is a closed formula with its weight, will be the intersection of the fuzzy sets $M(C_i)$, i.e.

$\mu_{M(\mathcal{H})}(I) = \min_{i=1,...,n} \mu_{M(C_i)}(I)$.

The *consistency* degree of $\mathcal{H}$ will be defined by $c(\mathcal{H}) = \max_I \mu_{M(\mathcal{H})}(I)$; it estimates the degree to which the set of models of $\mathcal{H}$ is not empty. The quantity Inc $(\mathcal{H}) = 1 - c(\mathcal{H})$ will be called degree of *inconsistency* of $\mathcal{H}$.

Finally we say that $\mathcal{F}$ is a *logical consequence* of $\mathcal{H}$ if and only if $\forall I, \mu_{M(\mathcal{F})}(I) \geq \mu_{M(\mathcal{H})}(I)$. Let us note that all these definitions subsume those of classical logic. We shall use the following notations:

- $\mathcal{H} \vdash C$ with $C = (C^* \, (N \, \alpha))$ if and only if from the set of necessity-valued clauses equivalent to $\mathcal{H}' = \mathcal{H} \cup \{\neg C \, (N \, 1)\}$ we can produce an $\alpha$-refutation (i.e. a deduction of $(\emptyset \, (N \, \alpha))$).
- $\mathcal{H} \models C$ if and only if C is a logical consequence of $\mathcal{H}$.

Then the following theorems hold:

<u>Theorem 1</u> : For any necessity-valued clauses $C = (C^* \, (N \, \alpha))$ and $C' = (C'^* \, (N \, \beta))$, $C, C' \vdash C''$ implies $C, C' \models C''$, where C" is a weighted clause.

<u>Proof</u> : we have to prove that $\forall I, \mu_{M(C'')}(I) \geq \mu_{M(C \wedge C')}(I)$.

- If I is a model of the classical formula $C^* \wedge C'^*$, then $\mu_{M(C \wedge C')}(I) = 1$ and the soundness of the classical resolution principle enables us to say that $I \in M(C''^*)$, where $C''^*$ is the resolvant of $(C^*, C'^*)$. Thus we have $\mu_{M(C'')}(I) = 1 \geq \mu_{M(C \wedge C')}(I)$.
- If I is not a model of $C^* \wedge C'^*$, then $\mu_{M(C \wedge C')}(I) \in \{1-\alpha, 1-\alpha'\}$, and $\mu_{M(C \wedge C')}(I) \leq \max(1-\alpha, 1-\alpha') = 1-\min(\alpha, \alpha')$. Besides, by definition we have $\mu_{M(C'')}(I) = 1-\min(\alpha, \alpha')$ and then $\mu_{M(C'')}(I) \geq \mu_{M(C \wedge C')}(I)$.  Q.E.D.

<u>Corollary</u> : let H be a set of necessity-valued clauses, then any necessity-valued clause C derived from $\mathcal{H}$ is a logical consequence of $\mathcal{H}$, i.e. the resolution principle for necessity-valued clauses is sound.

<u>Proof</u> : by induction on the refutation, using theorem 1.

<u>Theorem 2</u> : Let $\mathcal{H}$ be a set of necessity-valued clauses. If $\mathcal{H}$ is $\alpha$-inconsistent ($\alpha > 0$) then there is an $\alpha$-refutation from $\mathcal{H}$, i.e. the resolution principle for necessity-valued clauses is complete for the refutation.

<u>Sketch of the proof</u> : (the proof of the lemmas is omitted for the sake of brevity)

<u>Lemma 2.1</u> :
$\mathcal{H}$ is $\alpha$-inconsistent $\Leftrightarrow$ ($\emptyset$ (N $\alpha$)) is a logical consequence of $\mathcal{H}$.

Let us note $V_\alpha = (\emptyset$ (N $\alpha$)).

<u>Lemma 2.2.</u> :
$V_\alpha$ is a logical consequence of $\mathcal{H}$ ($\alpha > 0$) $\Rightarrow$ $V_\alpha$ is a logical consequence of $\mathcal{H}_\alpha$, where $\mathcal{H}_\alpha$ is the subset of the clauses of $\mathcal{H}$ whose valuations are greater than or equal to (N $\alpha$).

<u>Lemma 2.3.</u> :
$V_\alpha$ is a logical consequence of $\mathcal{H}_\alpha \Rightarrow \mathcal{H}_\alpha^*$ is inconsistent in classical logic, where $\mathcal{H}_\alpha^*$ is the set of clauses of $\mathcal{H}_\alpha$ without their weights.

<u>Lemma 2.4.</u> : ($\emptyset$ (N $\alpha$)) is a logical consequence of $\mathcal{H}_\alpha \Rightarrow$ there is a (N $\alpha$)-refutation from $\mathcal{H}$.

<u>Remark</u> : By contrast, let us notice that the probabilistic resolution principle is not complete. The resolution rule analogous to (1), for probabilistic logic (in terms of probability measures on a Boolean algebra of formulas, see Nilsson [18]), is

(5)
$$\begin{array}{l} \text{Prob}(p \vee q) \geq \alpha \\ \text{Prob}(\neg p \vee r) \geq \beta \\ \hline \text{Prob}(q \vee r) \geq \max(0, \alpha + \beta - 1) \end{array}$$

Let q and r be two ordinary propositions and let us take the following assignments

$$\begin{array}{ll} \text{Prob}(\neg r \vee q) \geq 0.7 & \text{(i)} \\ \text{Prob}(r) \geq 0.6 & \text{(ii)} \\ \text{Prob}(\neg q) \geq 0.5 & \text{(iii)} \end{array}$$

Now let us find the best lower bound of Prob($\neg q \vee r$). Let $\alpha, \beta, \gamma, \delta$ be respectively Prob($\neg q \wedge \neg r$), Prob($q \wedge \neg r$), Prob($\neg q \wedge r$) and Prob($q \wedge r$). The three last inequalities become

$$\begin{array}{ll} \alpha + \beta + \delta \geq 0.7 & \text{(j)} \\ \gamma + \delta \geq 0.6 & \text{(jj)} \\ \alpha + \gamma \geq 0.5 & \text{(jjj)} \end{array}$$
Prob($\neg q \vee r$) = $\alpha + \gamma + \delta$ = 1 - $\beta$ = 2 - (1 - ($\alpha + \beta + \delta$)) - (1 - ($\gamma + \delta$)) - (1 - ($\alpha + \gamma$)).

Using (j),(jj) and (jjj) we obtain Prob($\neg q \vee r$) $\geq 0.8$. The best lower bound of Prob($\neg q \vee r$) computed using (1),(2),(3) is 0.8 and by resolution, the best lower bound obtained (equal to the weight attached to the optimal refutation using pattern (5)) is only 0.6. Hence the resolution principle for probabilistic logic is not complete.

## 5. Extension to variable certainty weights

The lower bounds can be allowed to depend on variables involved in the clause. For instance

$$c(x) \quad (N \; \mu_P(x))$$

means that for any x, we are certain that c(x) is true at least with a necessity degree equal to $\mu_P(x)$. If $\mu_P$ is the characteristic function of a vague predicate P, it enables us to express that the more x satisfies P, the more certain the clause c(x). This would apply as well to a lower bound of a possibility measure. Note that the predicates involved in the clause c(x) are supposed to be non- vague. Although patterns (1) and (3) have been generalized to the case of vague predicates (see Dubois and Prade [7]), the clauses are supposed here to involve ordinary predicates only.

The instantiation of a variable certainty weight will be made with the instantiation of the involved variable, e.g. from ($\neg p(x) \vee q(x)$ (N $\varphi(x)$)) and (p(a) (N $\alpha$)) we infer (q(a) min($\alpha,\varphi(a)$)), or as soon the involved variable is eliminated : e.g., from ($\neg p(t) \vee q(y)$ (N $\varphi(y,t)$)) and (p(x) (N $\psi(x)$)) we can infer (q(y) (N min($\varphi(y,t), \psi(t)$))) for every value of t ; hence we infer the uncertain clause (q(y) (N $\sup_{t \in T}$ min($\varphi(y,t), \psi(t)$))) where T is the domain of the variable t.

Besides, it is possible to transform a predicate in a clause into a variable certainty weight and conversely. Indeed, suppose we have the certain clause C : ($\neg p(x) \vee q(x)$ (N 1)), then C is equivalent to C1 : ($\neg p(x)$ (N $\mu_{\neg Q}(x)$)) where $\mu_{\neg Q}$ is the characteristic function attached to the predicate $\neg q$, i.e. $\mu_{\neg Q}(x) = 1$ if x does not satisfy q, and $\mu_{\neg Q}(x) = 0$ if x satisfies q; thus C1 means that we are certain that x satisfies $\neg p$ as soon as it satisfies $\neg q$. Note that C1 can be obtained by resolution from C and the clause ($\neg q(x)$ (N $\mu_{\neg Q}(x)$)) which obviously holds. C is also equivalent to C2 : (q(x) (N $\mu_P(x)$)), and to C3 : ($\emptyset$ (N min($\mu_P(x), \mu_{\neg Q}(x)$))). It is easy to check that C1, C2 and C3 contain exactly the same information as C in the sense of the semantics introduced in section 4. For instance, C3 expresses that it is incoherent for an x to satisfy p and $\neg q$. Thus, variable certainty weights offer a kind of bridge between syntax and semantics. Similarly, from the uncertain clause C' : ($\neg p(x) \vee q(x)$ (N $\alpha$)) we can, in the same way, infer C'1 : $(\neg p(x)$ (N min $(\alpha, \mu_{\neg Q}(x))))$. Thus, each time we have not enough information for obtaining a refutation, or more generally each time we want to consider some predicates as hypotheses (i.e. the user may decide of their truth or falsity), we shall pass the subpart of the initial clause containing the involved predicates into the weight attached to the clause. Such a way of processing is close to the functionality of De Kleer's A.T.M.S. [2] since finding all the proof paths leading to a conclusion comes down to compute all environments where this conclusion holds. Each proof path corresponds to one environment that is captured by the final variable certainty weight.

### 6. Illustrative example

Let $\mathcal{H}$ be the following knowledge base :

1. If Bob attends a meeting, then Mary does not.
2. Bob comes to the meeting to-morrow.
3. Someone will come almost certainly to the meeting to-morrow, whose presence may (highly possibly, but not certainly at all) make the meeting not quiet.
4. If Albert comes to-morrow and Mary does not, then it is almost certain that the meeting will not be quiet.
5. It is likely that Mary or John will come to-morrow.
6. If John comes to-morrow, it is rather likely that Albert will come.
7. The later John will arrive to the meeting to-morrow, the more certain we are that the meeting will be quiet, and if he does not come at all, it is certain that the meeting will be quiet.

We note that the third sentence is translated into two clauses, C3 and C4, by introducing the Skolem constant "a" . The first part of the 7$^{th}$ sentence will be expressed in possibilistic logic by a clause whose valuation depends on the time John will arrive; we will denote by $\mu_{late}(t)$ the membership function of the fuzzy set "late", pictured on Figure 1. The 7$^{th}$ sentence is represented by means of clause C8 and C9. Clause C10 expresses the relationship between "arrives" and "comes".

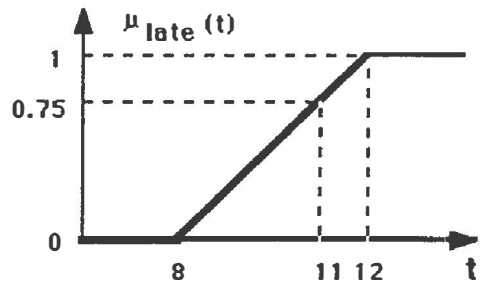

Figure 1

The knowledge base can be represented by the following weighted clauses :

| C1 | $\neg$comes(Bob,x) $\vee$ $\neg$comes(Mary,x) | (N 1) |
| C2 | comes(Bob,m) | (N 1) |
| C3 | $\neg$comes (a,m) $\vee$ $\neg$quiet(m) | ($\Pi$ 0.8) |
| C4 | comes (a,m) | (N 0.7) |
| C5 | comes(Mary,m) $\vee$ $\neg$comes(Albert,m) $\vee$ $\neg$quiet(m) | (N 0.8) |
| C6 | comes(John,m) $\vee$ comes(Mary,m) | (N 0.7) |
| C7 | comes(John,m) $\vee$ comes(Albert,m) | (N 0.6) |
| C8 | $\neg$arrives(John,m,t) $\vee$ quiet(m) | (N $\mu_{late}(t)$) |
| C9 | comes(John,m) $\vee$ quiet (m) | (N 1) |
| C10 | $\neg$arrives(x,y,z) $\vee$ comes(x,y) | (N 1) |

If we want try to prove that the meeting to-morrow will not be quiet, we add the clause C0 : quiet(m) (N 1). Then it can be checked that there exist two possible refutations, represented on figures 2 and 3.The second refutation is the optimal one. We proved that N($\neg$ quiet(m))≥0.6, i.e. it is rather likely that the meeting to-morrow will not be quiet.

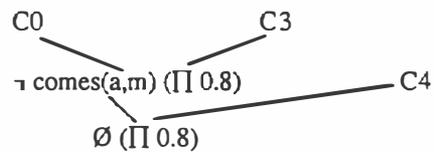

Figure 2

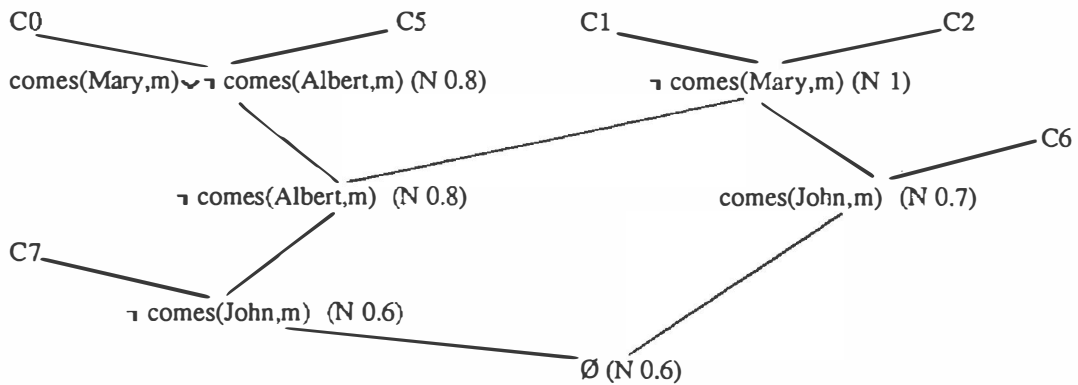

Figure 3

85

Suppose we want to consider comes (Bob,x) as an hypothesis (in the sense of the section 4) ; then C2 disappears and C1 becomes (¬comes(Mary,x) (N $\mu_{comes(Bob)}(x)$)) ; the two refutations lead u to conclude that the meeting tomorrow will not be quiet with a weight equal to max(($\Pi$ 0.8), (N min(0.6, $\mu_{comes(Bob)}(m)$))) ; then, if we choose not to fix comes(Bob,m) to true or false, or if we fix it to false, then we can only deduce $\Pi(\neg quiet(m)) \geq 0.8$ ; whereas if we fix it to true, we can deduce $N(\neg quiet(m)) \geq 0.6$.

## 7. Updating uncertain knowledge bases

It is possible to work with a partially inconsistent knowledge base $\mathcal{H}$. Let $\alpha$ be the inconsistency degree of $\mathcal{H}$. From $\mathcal{H}$ we want to prove c with some necessity degree. Let $\beta$ be the necessity degree of the optimal empty clause derived from $\mathcal{H}' = \mathcal{H} \cup \{(\neg c\ (N\ 1))\}$. We always have $\beta \geq \alpha$. More precisely, we have min $(\beta, \beta') = \alpha$ where $\beta'$ is the degree attached to the optimal empty clause derived from $\mathcal{H} \cup \{(c\ (N\ 1))\}$. Having min $(\beta, \beta') > 0$ may seem contradictory with possibility theory where $min(N(p), N(\neg p)) = 0$ holds for any p. But we must not forget that this situation happens only if the knowledge base is (partially) inconsistent, i.e. violates the axioms of possibility theory; moreover, at most one of the two conclusions p or ¬p will be taken as valid.

Noticeably, if $\beta > \alpha$, there exists a consistent sub-base $\mathcal{S}$ of $\mathcal{H}$ from which we can infer (c (N $\beta$)) by resolution, and then we will consider the proof of (c (N $\beta$)) as valid. Indeed, the $\beta$-refutation uses only clauses with a necessity degree greater or equal to $\beta$, i.e. strictly greater than $\alpha$ ; $\mathcal{S}$ is the subset of $\mathcal{H}$ containing the clauses of $\mathcal{H}$ used in this $\beta$-refutation. We cannot produce any refutation from $\mathcal{S}$ only, since this refutation would have a valuation strictly greater to (N $\alpha$), and then $\mathcal{H}$ would have an inconsistency degree greater than $\alpha$.

Moreover, adding to a consistent knowledge base $\mathcal{H}$, a clause (c (N $\alpha$)) that makes it partially inconsistent, produces what is similar to a non-monotonic behavior. Namely, if from $\mathcal{H}$ a conclusion (p (N $\beta$)) can be obtained by refutation, it may happen that from $\mathcal{H}' = \mathcal{H} \cup \{(c\ (N\ \alpha))\}$, an opposite conclusion ($\neg p$ (N $\gamma$)) with $\gamma > 1 - c(\mathcal{H}') \geq \beta$ can be derived, where $1 - c(\mathcal{H}')$ is the degree of inconsistency of $\mathcal{H}'$.

Example : we work again with the knowledge base of Sec.6. Suppose we add to $\mathcal{H}$ the clause (arrives(John,m,11) (N 1)) i.e. $\alpha = 1$, expressing that we are now certain that John comes to-morrow at 11 o'clock. Let $\mathcal{H}'$ be the new knowledge base. The inconsistency degree of $\mathcal{H}'$ is 0.6, i.e. $1-c(\mathcal{H}') = 0.6$.

Now the proof of (¬quiet(m) (N 0.6)) (it corresponds to $\beta = 0.6$) is no longer valid ; but we can prove (quiet(m) (N 0.75)), i.e. $\gamma = 0.75$, since $\mu_{late}(11) = 0.75$ :

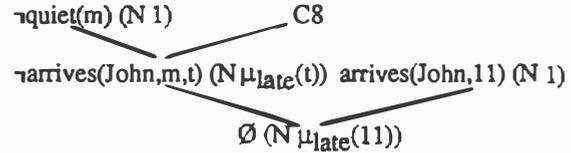

using only a consistent part of $\mathcal{H} \cup \{(arrives(John,m,11)\ (N\ 1))\}$. Thus a non-monotonic behaviour can be captured in this framework.

Indeed our approach to partially inconsistent knowledge consists in selecting a particular extension (in the sense of default logic) on the basis of the certainty weights attached to the clauses. For example, if we work with the partially inconsistent base $\mathcal{S} = \{(u\ (N\ \alpha)), (\neg u \vee v\ (N\ \beta)), (\neg v\ (N\ \gamma))\}$, we will prefer the extension

- $\{\neg u, \neg v\}$ if $min(\alpha, \beta, \gamma) = \alpha$
- $\{u, \neg v\}$ if $min(\alpha, \beta, \gamma) = \beta$
- $\{u, v\}$ if $min(\alpha, \beta, \gamma) = \gamma$

Thus, if $\alpha > \beta > \gamma$, we will prefer $\{u, v\}$ to $\{u, \neg v\}$ and $\{u, \neg v\}$ to $\{\neg u, \neg v\}$. The order on the valuations leads to a partial ordering on the consistent sub-bases of $\mathcal{S}$.

It becomes clear from this paper that what is important is more the ordering among the possibility or necessity degrees than their precise value. See Dubois [3] which discusses qualitative possibility measures (defined in terms of a relation "at least as possible as"), as well as qualitative necessity measures, as natural counterparts of numerical possibility and necessity measures. It can be checked that the required properties of the relation underlying a qualitative necessity measure are strictly equivalent to the properties characterizing a relation of epistemic entrenchment in the sense of Gärdenfors [11]. See Dubois, Prade [9]. Thus the updating process offered by the possibilistic framework may appear less surprising, as being in agreement with more general theories of revision.

## 8. Concluding remarks

Possibilistic logic offers a convenient and rigorous framework for handling uncertainty in automated deduction. Possibility theory enables us to represent states of partial ignorance. The results are only sensitive to the ordering of numbers, which is quite in agreement with a rather poor quality of the available knowledge about certainty levels. Besides the use of variable certainty weights enables us to accommodate hypothetical reasoning.

The resolution patterns and the refutation strategy presented in this paper are implemented in the system POSLOG (for POSsibilistic LOGic) on a micro-computer.

## References

[1] Baldwin J.F. , Evidential support logic programming, Fuzzy Sets and Systems 24 (1987), 1-26.
[2] De Kleer J. , An assumption-based truth maintenance system. Artificial Intelligence 28, 127-162.
[3] Dubois D. , Belief structures, possibility theory and decomposable confidence measures on finite sets. Computers and Artificial Intelligence 5(1986), 403-416.
[4] Dubois D., Lang J., Prade H., Theorem proving under uncertainty - A possibility theory-based approach. Proc. 10th Inter. Joint Conf. on Artificial Intelligence, Milano, Aug., 1987, 984-986.
[5] Dubois D., Prade H. (with the collaboration of Farreny H., Martin- Clouaire R., Testemale C.) Possibility Theory. An Approach to Computerized Processing of Uncertainty. Plenum Press, New York, 1988.
[6] Dubois D., Prade H. Necessity measures and the resolution principle. IEEE Trans. Systems, Man and Cybernetics, Vol. 17 (1987), 474-478.
[7] Dubois D., Prade H. Resolution principles in possibilistic logic. To appear in Inter. J. of Approximate Reasoning. In Tech. Rep. LSI n° 257, Feb. 1988.
[8] Dubois D., Prade H., An introduction to possibilistic and fuzzy logics. In Non standard logics for automated reasoning (P.Smets et al., Eds), Academic Press, New York, 287-326.
[9] Dubois D., Prade H. (1989), Epistemic entrenchment, partial inconsistency and abnormality in possibilistic logic. BUSEFAL n° 39, 66-73, (Lab. LSI-IRIT, Univ. P. Sabatier, Toulouse)
[10] Dubois D., Prade H., Testemale C. In search of a modal system for possibility theory. 8th Europ. Conf. on Artificial Intelligence, 501-506, Münich, August 1988.
[11] Gärdenfors P. (1988), Knowledge in Flux. Modeling the Dynamics of Epistemic States. MIT Press, Cambridge, Ma.
[12] Grosof B., Non-monotonicity in probabilistic reasoning, in Uncertainty in Artificial Intelligence 2, J.F Lemmer and L.N. Kanal (editors), North-Holland, Amsterdam, 1988, 237-249.
[13] Hinde C.J., Fuzzy Prolog, Int. J. Man-Machine Studies 24 (1986), 659-695.
[14] Ishizuka M, Kanai N., Prolog-Elf incorporating fuzzy logic, Proc. 9th Inter. Joint Conf. on Artificial Intelligence, Los Angeles, August 1985, 701-703.
[15] Léa Sombé (1989) (alias P. Besnard, M.O. Cordier, D. Dubois, L. Fariñas del Cerro, C. Froidevaux, Y. Moinard, H. Prade, C. Schwind, P. Siegel), Raisonnements sur des Informations Incomplètes en Intelligence Artificielle, Teknéa, Marseille. Also in Revue d'Intelligence Artificielle, Hermès, Vol. 2 (3-4), 9-210.
[16] Lee R.C.T. Fuzzy logic and the resolution principle, J. of the Assoc. for Computing Machinery, 19 (1972), 109-119.
[17] Martin T.P., Baldwin J.F., Pilsworth B.W., The implementation of FProlog - A fuzzy Prolog interpreter, Fuzzy Sets and Systems 23 (1987), 119-129.
[18] Nilsson N.J., Probabilistic Logic, Artificial Intelligence Vol. 28(1986), 71-88.
[19] Orci I.P., Programming in Possibilistic Logic, Research Report UMINF-137.87A, Departement of Computing Science, Umeå University, Sweden, June 1988. To appear in International Journal of Expert Systems: Research and Applications 2(1), 1989.
[20] Shapiro E.Y., Logic programs with uncertainties: a tool for implemented rule-based systems, Proc 8th Int. Joint Conf. on Artificial Intelligence, Karlsruhe (1983), 529-532.
[21] Shen Z., Ding L., Mukaidono M., A theoretical framework for the Fuzzy Prolog machine, in Fuzzy Computing, M.M. Gupka and T. Yamakawa (Eds), North-Holland, Amsterdam,89-100, 1988.
[22] Umano M., Fuzzy-set Prolog, Proc. 2nd Inter. Fuzzy Systems Association Congress, Tokyo, 750-753, 1987.
[23] Van Emden M.H., Quantatitative deduction and its fixpoint theory, J. Logic Programming, 1, 37-53, 1986.
[24] Zadeh L.A. Fuzzy sets as a basis for a theory of possibility. Fuzzy Sets and Systems, 1, 1978, 3-28.